\documentclass[sigconf]{acmart}
\AtBeginDocument{%
  \providecommand\BibTeX{{%
    \normalfont B\kern-0.5em{\scshape i\kern-0.25em b}\kern-0.8em\TeX}}}

\setcopyright{acmlicensed}
\copyrightyear{2022}
\acmYear{2022}
\setcopyright{acmlicensed}\acmConference[GECCO '22]{Genetic and
Evolutionary Computation Conference}{July 9--13, 2022}{Boston, MA, USA}
\acmBooktitle{Genetic and Evolutionary Computation Conference (GECCO '22),
July 9--13, 2022, Boston, MA, USA}
\acmPrice{15.00}
\acmDOI{10.1145/3512290.3528823}
\acmISBN{978-1-4503-9237-2/22/07}
\acmPrice{15.00}

\usepackage[ruled]{algorithm2e}
\usepackage{algpseudocode}

\usepackage{xspace}
\usepackage{units}

\newcommand{\mymath}[1]{\ensuremath{#1}\xspace}
\newcommand{\reals}{\mymath{\mathbb R}}

\definecolor{myred}{rgb}{0.8,0,0}
\definecolor{mygreen}{rgb}{0,0.6,0}
\definecolor{myblue}{rgb}{0,0,0.7}

\newcommand{\spea}{{\sc spea2}\xspace}
\newcommand{\nsga}{{\sc nsga-ii}\xspace}
\newcommand{\mome}{{\sc mome}\xspace}
\newcommand{\me}{{\sc map-elites}\xspace}
\newcommand{\pgame}{{\sc pga-map-elites}\xspace}
\newcommand{\cmame}{{\sc cma-map-elites}\xspace}
\newcommand{\moqd}{{\sc moqd}\xspace}
\newcommand{\qd}{{\sc qd}\xspace}

\begin{document}

\title{Multi-Objective Quality Diversity Optimization}


\author{Thomas Pierrot}
\authornote{Both authors contributed equally to this research.}
\affiliation{%
  \institution{InstaDeep}
  \streetaddress{30, rue Le Peletier}
  \city{Paris}
  \country{France}
}
\email{t.pierrot@instadeep.com}

\author{Guillaume Richard}
\authornotemark[1]
\affiliation{%
  \institution{InstaDeep}
  \streetaddress{30, rue Le Peletier}
  \city{Paris}
  \country{France}
}
\email{g.richard@instadeep.com}

\author{Karim Beguir}
\affiliation{%
  \institution{InstaDeep}
  \streetaddress{30, rue Le Peletier}
  \city{Paris}
  \country{France}
}
\email{kb@instadeep.com}

\author{Antoine Cully}
\affiliation{%
  \institution{Imperial College London}
  \city{London}
  \country{United Kingdom}}
\email{a.cully@imperial.ac.uk}

\renewcommand{\shortauthors}{Pierrot and Richard, et al.}

\begin{abstract}
In this work, we consider the problem of Quality-Diversity (QD) optimization with multiple objectives. QD algorithms have been proposed to search for a large collection of both diverse and high-performing solutions instead of a single set of local optima. Searching for diversity was shown to be useful in many industrial and robotics applications. On the other hand, most real-life problems exhibit several potentially conflicting objectives to be optimized. Hence being able to optimize for multiple objectives with an appropriate technique while searching for diversity is important to many fields. Here, we propose an extension of the \me algorithm in the multi-objective setting: Multi-Objective \me (\mome). Namely, it combines the diversity inherited from the \me grid algorithm with the strength of multi-objective optimizations by filling each cell with a Pareto Front. As such, it allows to extract diverse solutions in the descriptor space while exploring different compromises between objectives. We evaluate our method on several tasks, from standard optimization problems to robotics simulations. Our experimental evaluation shows the ability of \mome to provide diverse solutions while providing global performances similar to standard multi-objective algorithms.

\end{abstract}



\begin{teaserfigure}
  \includegraphics[width=0.85\textwidth]{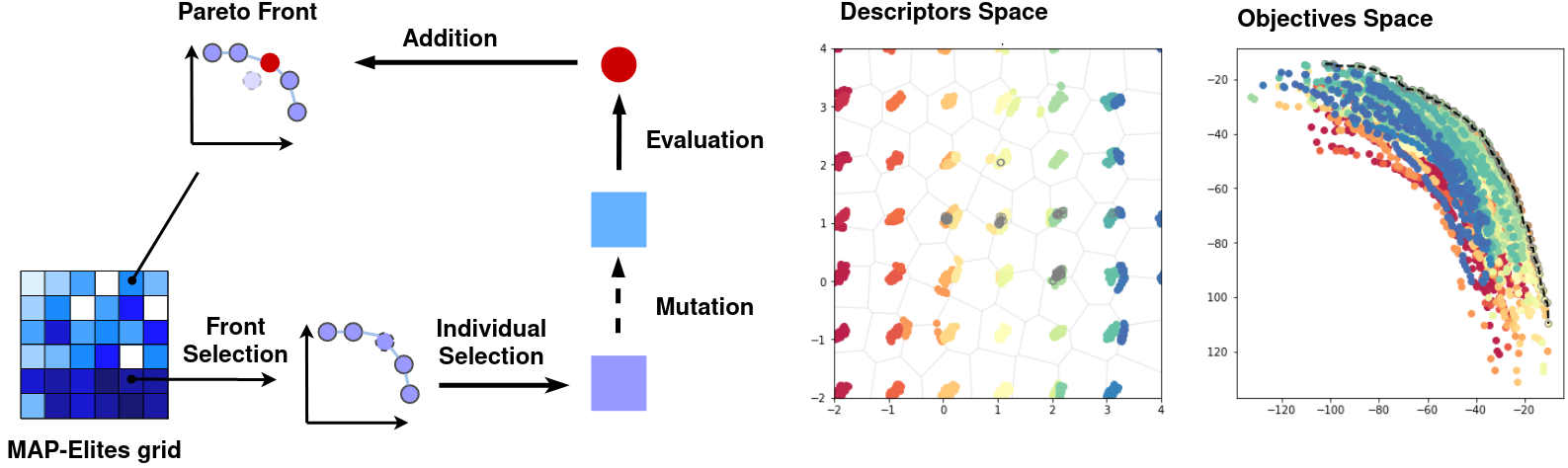}
  \caption{On the left panel, high-level concepts of \mome. We extend the \qd objective to the multi-objective optimization setting, where maximizing the fitness is replaced by maximizing a Pareto front hyper-volume in each niche of the descriptor space. To do so, \mome evolves a local Pareto front in each cell. On the right panel, an example of the descriptor space coverage and objective space coverage obtained with \mome on the Rastrigin\_proj task. \mome finds a collection of performing Pareto fronts, spanning the descriptor space, that can also be combined to obtain a single highly performing global Pareto front. Interestingly, we find that several cells contribute to it showing the benefits of covering a descriptor space for multi-objective optimization.}
  \label{fig:teaser}
\end{teaserfigure}

\maketitle

\section{Introduction}

Quality-Diversity (QD) Optimization algorithms \cite{cully2017quality, chatzilygeroudis2021quality} have changed the classical paradigm of optimization: inspired by natural evolution, the essence of QD methods is to provide a large and diverse set of high-performing solutions rather than only the best one. This core idea is crucial to many applications, from robotics where it allows robots to adapt to unexpected scenarios \cite{cully2015robots} to game design \cite{khalifa2018talakat} or investment strategies \cite{zhang2020autoalpha}. This becomes particularly interesting when the optimization process is done using a surrogate model instead of the real one as the highest performing solution according to the simulation might not transfer well in reality.

Quality Diversity algorithms tackle this issue by not only optimizing the fitness but also aiming to cover a variety of user-defined features of interest. Novelty Search \cite{lehman2011abandoning, lehman2011evolving} was the first method to explicitly promote diversity by optimizing the novelty of a solution. Building on this paradigm, illumination algorithms \cite{mouret2015illuminating} have been developed with a core principle: "illuminating" the search space by creating local descriptor niches where the solution with the highest fitness is extracted. Namely, MAP-Elites \cite{mouret2015illuminating} explicitly divides the descriptor space in niches by defining a grid \cite{vassiliades2016scaling} and each niche stores its best solution. The method has been further extended to tackle Multi-Task Optimization \cite{mouret2020quality}, efficiently evolve large neural network with Policy Gradient \cite{nilsson2021policy} or be robust to noise \cite{flageat2020fast}. \me has also been used in a multi-objective scenario where one grid per objective is maintained .

On the other hand, multi-objective optimization \cite{deb2014multi} is a very popular and useful
field of research and application. Indeed, most applications involve multiple conflicting objectives and it is key to have a view of the different possible trade-offs to make a decision. For instance, training a robot to maximize its forward speed while minimizing the energy it uses involves two conflicting terms. While it is possible to combine the objectives into one, multi-objective approaches look for a set of Pareto-optimal solutions, \textit{ie} none of which can be considered better than another when every objective is of importance.

Traditional multi-objective optimization methods are designed to generate a set of solutions that approximate the set of all optimal trade-offs, called the Pareto Front. Evolutionary algorithms are a natural approach and numerous methods have been proposed, differing on their underlying selection scheme: Non-dominated Sorted Genetic Algorithm (\nsga) \cite{deb2002fast}, Strength Pareto Evolutionary Algorithm (\spea) \cite{zitzler2001spea2} or multi-objective evolutionary algorithm based on decomposition (MOEA/D) \cite{zhang2007moea} are examples of such approaches. Those methods generally allow to find high-performing solutions by insuring diversity in the objective space. However, to the best of our knowledge, no method explicitly tackles the issue of Multi-Objective Quality Diversity. A method of interest was introduced with Pareto Gamuts \cite{makatura2021pareto} where authors tackle multi-objective design problems under different contexts, where contexts are defined as different values of a real-valued variable. In this work, we consider more general descriptors in our search for diversity. 

In this work, we propose to design a novel method for Multi-Objective Quality Diversity (\moqd) optimization. Namely, we introduce Multi-Objective MAP-Elites (\mome) which divides the descriptor space in niches with a tessellation method and illuminates each cell of the tessellation by filling it with its set of Pareto optimal solutions. As such, it allows to find solutions that span the space of descriptors while being locally Pareto efficient. We evaluate \mome on four different tasks: a traditional real-valued optimization problem taken from the literature and three robotic control tasks using a physical simulator. We compare \mome against standard baselines and show that our approach not only helps to find more diverse solutions but that diversity in the descriptors space sometimes helps to find better solutions in terms of Pareto efficiency. Our implementation is publicly available and fully done in Python using the high performing Jax \cite{jax2018github} library which has shown to significantly accelerate QD applications~\cite{Lim2022Accelerated}. We believe this will provide very useful to build upon for further advancements in \moqd optimization. 

\section{Problem Formulation}

\subsection{Quality-Diversity Optimization}
We now introduce our framework for Multi-Objective Quality Diversity (\moqd), adapted from a recent work \cite{fontaine2021differentiable}. The Quality Diversity (\qd) problem assumes a single objective function $\mathcal{X} \rightarrow \reals$, where $\mathcal{X}$ is called search space, and $d$ descriptors $c_i: \mathcal{X} \rightarrow \reals$, or as a single descriptor function $\mathbf{c}: \mathcal{X} \rightarrow \reals^d$. We note $S = \mathbf{c}(\mathcal{X})$ the descriptor space formed by the range of $\mathbf{c}$.

\qd algorithms of the family of the \me algorithm, discretize the descriptor space $S$ via a tessellation method. Let $\mathcal{T}$ be the tessellation of $S$ into $d$ cells $S_i$. The goal of \qd methods is to find a set of solutions $\mathbf{x}_i \in \mathcal{X}$ so that each solution $\mathbf{x}_i$ occupies a different cell $S_i$ in $\mathcal{T}$ and maximizes the objective function within that cell. The \qd objective can thus be formalized as follows:

\begin{align}
    \max\limits_{\mathbf{x} \in \mathcal{X}} \sum\limits_{i=1}^d f(\mathbf{x}_i), \ \text{where} \ \forall i, \ \mathbf{c}(\mathbf{x}_i) \in S_i. 
\end{align}

\subsection{Multi-Objective Optimization}

We consider an optimization problem over the search space $\mathcal{X}$ with $k$ associated objective functions $f_i : \mathcal{X} \rightarrow \reals$, or as a single function $\mathbf{f}: \mathcal{X} \rightarrow \reals^k$. In multi-objective optimization, the goal is to find a solution $\mathbf{x} \in \mathcal{X}$ that maximizes at once the different objectives. The multi-objective problem can be formalized as

\begin{align}
    \max\limits_{\mathbf{x} \in \mathcal{X}}{\left(f_1(\mathbf{x}), ..., f_k(\mathbf{x})\right)}.
\end{align}

In most problems, there does not exist any solution that concurrently maximizes every objective. In fact, objectives are often antagonist which implies that maximizing one objective alone can result in minimizing the others. Comparing solutions in a multi-objective setting is challenging too. In the mono-objective \me algorithm, solutions are replaced in their cell if another solution with higher value for the single objective is found. 

To compare solutions in the multi-objective setting, we rely on the notion of Pareto domination. Given two solution candidates $\mathbf{x}_1, \mathbf{x}_2 \in \mathcal{X}$, $\mathbf{x}_1$ Pareto dominates $\mathbf{x}_2$, noted $x_1  \succ x_2$, if $\forall 1 \leq i \leq k, f_i(\mathbf{x}_1) > f_i(\mathbf{x}_2)$.

Given a set of candidate solutions $\mathcal{S} \in \mathcal{X}$, we define the Pareto front over this set, $\mathcal{P}(\mathcal{S})$, as the subset of solutions in $\mathcal{S}$ that are not Pareto dominated by any other solution in $\mathcal{S}$. In other words, the Pareto front over a set of solutions corresponds to all the best possible trade-offs with respect to the set of objectives, there is no solution in the Pareto front that is better that the others in the front on all the objectives at the same time. Canonical multi-objective optimization algorithms aim to find the Pareto front of solutions over the search space $\mathcal{X}$ that we call Optimal Pareto front.

In the mono-objective optimization setting, optimization methods can be compared in terms of the best value found for the objective function. In the multi-objective setting, the optimization methods we consider return a Pareto front of solutions. Thus, to compare them we need metrics to compare Pareto fronts. One simple and robust way to compare two Pareto fronts is to compare their \textit{hyper-volumes} \cite{hypervolume}. The hyper-volume $\Xi$ of a Pareto front $\mathcal{P}$ requires the definition by the user of a reference point $\mathbf{r} \in \reals^K$. It is defined as the Lebesgue measure of the set of points dominated by $\mathcal{P}$ \cite{zitzler1999multiobjective}:

\begin{align}
\Xi(\mathcal{P}) = \Gamma (\{x \in \mathcal{X} | \exists s \in \mathcal{P},  s \succ x \succ r  \} )
\end{align}

\begin{figure}
    \centering
    \includegraphics[width=0.45\textwidth]{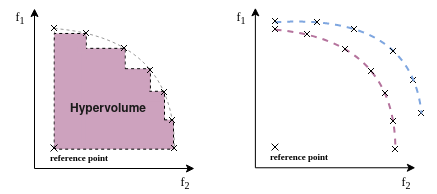}
    \caption{On the left panel, an example of a Pareto front hypervolume computation when there are $K=2$ objective functions. On the right panel, an example of two Pareto fronts when $K=2$, where the front in blue strictly dominates, in terms of hypervolume, the front in purple given the showed reference point.}
    \label{fig:hypervolume}
\end{figure}

As described in Figure~\ref{fig:hypervolume}, in the case of two objectives, this metric simply corresponds to the area between the curve and the reference point. When comparing Pareto fronts for a given optimization problem, the reference point is fixed in advance and remains unchanged for all hyper-volumes computations. The hypervolume indicator induces an order on Pareto Fronts which is monotonic and invariant to scale \cite{knowles2002local, zitzler1999evolutionary}, the main drawbacks being its dependence on the definition of the reference point and its computation cost in high dimension \cite{brockhoff2010optimal}. 

\subsection{Multi-Objective Quality Diversity Optimization}
\label{sec:moqd}
In this work, we introduce the Multi-Objective Quality Diversity (\moqd) problem that assumes $k$ objective functions $f_i: \mathcal{X} \rightarrow \reals$, or as a single function $\mathbf{f}: \mathcal{X} \rightarrow \reals^k$ and $d$ descriptors $c_i: \mathcal{X} \rightarrow \reals$, or as a single descriptors function $\mathbf{c}: \mathcal{X} \rightarrow \reals^d$.

As in the \qd framework, we assume that the descriptor space $S$ is discretized via a tessellation method into $d$ cells $S_i$. In the mono-objective framework, a commonly accepted metric is the \qd-score, defined as the sum of the fitnesses of the solution in each cell. The goal of \moqd methods is to find in each cell, a Pareto front of solutions $\mathcal{P}(S_i)$ that maximizes its hyper-volume. In other word, for each niche of the descriptor space we seek for all the best possible trade-offs regarding the objective functions $\mathbf{f}$. Following the \qd-score \cite{pugh2015confronting}, \moqd objective can be formalized as follow:

\begin{align}
\max\limits_{\mathbf{x} \in \mathcal{X}} \sum\limits_{i=1}^d \Xi(\mathcal{P}_i), \ \text{where} \ \forall i, \ \mathcal{P}_i = \mathcal{P}(\{\mathbf{x} \ | \ \mathbf{c}(\mathbf{x}) \in S_i \}). 
\end{align}

In the next sections, we will refer to this objective as the \moqd score. While our goal is to build methods that maximize the \moqd score, i.e. methods that find $d$ high-performing Pareto fronts, we may also, for comparison purpose, study the Pareto front over all the solutions found by the algorithm regardless of their descriptor which we refer to as the global Pareto front. In practice, we found that in many cases the global Pareto front does not correspond exactly to the Pareto front of one of the tessellation cells but is rather constituted by solutions coming from several niches. Thus, the global Pareto front may have an hyper-volume larger than the largest hyper-volume over the cells Pareto fronts.

\section{Background and Related Works}

\subsection{MAP-Elites}

Quality diversity algorithms aim to produce a large collection of solutions as diverse and high-performing as possible. MAP-Elites is a \qd method that discretizes the space of possible descriptors (BD) into a grid (also called archive) via a tessellation method. Its goal is to fill each cell of this grid with the highest performing individuals. To do so, MAP-Elites firstly initializes a grid over the BD space and secondly initializes random solutions, evaluates them and inserts them in the grid. Then successive iterations are performed: (i) select and copy solutions uniformly over the grid (ii) mutate the copies to create new solution candidates (iii) evaluate the candidates to determine their descriptor and objective value (iv) find the corresponding cells in the tessellation (v) if the cell is empty, introduce the candidate and otherwise replace the solution already in the cell by the new candidate if it has a greater objective function value. Competition between candidates and solutions of the same cells improves the quality of the collection, while filling additional cells increases its diversity.

% The closest idea to Map-Elites is Novelty Search \cite{lehman2011abandoning} which aims to find "novel" solutions in the input space, which can be extended to a descriptor space. Novelty Search has been extended to incorporate a fitness objective (Novelty Search with Local Competition \cite{lehman2011evolving}) and recently used as additional objective to a mono-objective optimization problem \cite{mouret2011novelty}. 

\subsection{Evolutionary multi-objective optimization}

Evolutionary Algorithms (EA) have attracted early attention for multi-objective optimization. They all start with a random population and iterate using mutation and selection. The different EA differ in their selection process. Here we briefly describe two baseline methods: Non-dominated Sorted Genetic Algorithm (NSGA-II) \cite{deb2002fast} and Strength Pareto Evolutionary Algorithm (\spea) \cite{zitzler2001spea2}.\\

\textbf{NSGA-II} is an evolutionary algorithm designed for optimization with multiple objectives. The selection of the new generation’s members of \nsga is made of two main parts: a non-dominated sorting and a crowded comparison. In the non-dominated sorting, the candidates are ranked according to their non-dominance: every member of the Pareto front is non-dominated and has a domination counter of $0$, then every member dominated only by members of the Pareto front have a domination counter of $1$ etc. Candidates are added to the archive based on their domination counter until it is impossible to add every candidate of the same domination counter without exceeding the maximum number of solutions. Then candidates of the same domination counter are ranked using crowding distances, which are an estimate of the distance between a candidate and its neighbours in the objective space. Candidates in sparsely populated areas are favoured.\\

\textbf{SPEA2} is also an evolutionary algorithm. Building over {\sc spea}, \citet{zitzler2001spea2} proposed to simplify it without decreasing performances. In the selection process, each candidate is ranked according to the number of other candidates dominating it. Hence, a candidate in the Pareto Front has a Strength Pareto score of $0$, while a candidate dominated by $2$ solutions has a Strength Pareto score of $2$. Similarly to \nsga, candidates are added to the archive based on this score until it is impossible to add more. Then candidates with the same score are sorted using a density estimation in the objective space based on $K$-neighbours, and candidates in sparsely populated ares of the objective space are favoured.\\

Those two methods favour the exploration in the objective space but do not consider a descriptor space. Subsequent methods, such as MOEA/D \cite{zhang2007moea} or MOPSO \cite{coello2002mopso} were designed with the same goal. Two other methods took interest in novelty search as a tool to improve multi-objective optimization: Novelty-based Multiobjectivization \cite{mouret2011novelty} considers novelty as an additional objective while MONA \cite{vargas2015general} uses novelty in a differential evolution algorithm as a criterion in the selection process. Both methods showed that for some applications, novelty search helps to find higher performing solutions that would not be found with method optimizing only for quality. However, to the best of our knowledge, no method has been designed specifically for Multi-Objective Quality Diversity.

\section{Multi-Objective MAP-Elites (MOME)}

In this section, we introduce a novel algorithm dubbed Multi-Objective MAP-Elites (\mome) that extends the \me algorithm to solve the \moqd problem.

The idea behind \mome is to progressively cover the descriptor space while building Elite Pareto fronts, i.e. Pareto fronts that maximize the hyper-volume in each niche. In opposition to \me that maintains either none or a single solution in each cell, \mome stores Pareto fronts of solutions in each cell. In order to benefit from the acceleration offered by compiled linear algebra libraries such as Jax, the \mome grid and all the applied operations are vectorized. Therefore, the Pareto fronts are assumed to have a maximum size $P$. If this size is reached, solutions are replaced randomly in the front. This same technique is used for the implementation of all the considered baselines that compute Pareto fronts. 

The general outline of \mome is presented in Figure~\ref{fig:teaser}. Following \me, to ensure an unbiased coverage of the descriptor space, a batch of cells is sampled uniformly with replacement in the grid at the beginning of each iteration, each cell containing a Pareto front. One solution is then uniformly sampled in each Pareto front and copied. As usual in evolutionary methods, the batch of solutions is then mutated and crossed-over using mutation and cross-over operators $f_{\text{mut}}: \mathcal{X} \rightarrow \mathcal{X}$ and $f_{\text{cross}}: \mathcal{X} \times \mathcal{X} \rightarrow \mathcal{X}$. 

The obtained offspring are then scored to compute their objectives and descriptors and inserted in the cell corresponding to its descriptors and all the Pareto front are updated. When adding an offspring to a Pareto front, we distinguish two cases:

\begin{itemize}
    \item There is one solution in the front that strictly dominates the offspring, in this case the offspring is dropped.
    \item There is no solution in the front that strictly dominates the offspring, in this case the offspring is added to the front. If the offspring were to strictly dominate solutions in the front, these solutions are dropped.
\end{itemize}

This insertion mechanism ensures that the set of solutions in each cell always forms a Pareto front. We point out that given this addition rule, in opposition to standard \me, the number of solutions in the grid may decrease from one iteration to another. Nonetheless, we observe in our experiments that this number usually increases on average until filling the grid in our experiments.

This cycle is repeated for a budget of $N$ iterations. The algorithm is warm-started with an initial population of solutions, sampled randomly in $\mathcal{X}$ and inserted in the grid following the addition rules detailed above. See Algorithm~\ref{alg:pseudocode} for the pseudo-code of \mome.

\makeatletter
\newcommand{\removelatexerror}{\let\@latex@error\@gobble}
\makeatother

\begin{figure}[h!]
\removelatexerror
\centering
\resizebox{0.45\textwidth}{!}{%
\begin{algorithm}[H]
    \small
    \SetAlgoLined
    \DontPrintSemicolon
    \SetKwInput{KwInput}{Given}
    \KwInput{
    \begin{itemize}
        \item the number of cells $M$ and the maximum Pareto front length $P$
        \item the batch size $B$ and the number of iterations $N$
        \item the descriptors function $\mathbf{m}$ and the multi-objective function $\mathbf{f}$
        \item the mutation and crossover operators $h_{\text{mut}}$ and $h_{\text{cross}}$
        \item the mutation proportion $p_{\text{mut}}$
        \item the initial population of solution candidates $\{\mathbf{x}_k\}$
    \end{itemize}
    }
    \texttt{\\}
    
    \tcp{Initialization}
    Use CVT to divide the descriptor space into $M$ Voronoi cells\;
    In each cell initialize an empty Pareto front of size $P$\;
    For each initial solution, find the cell corresponding to its descriptor\;
    Add initial solutions to their cells and recompute Pareto fronts\;
    \texttt{\\}
    
    \tcp{Main loop}
    $n_{\text{steps}} \leftarrow 0$\;
    \While{$n_{\text{steps}} < N$}{
    \texttt{\\}
    
        \tcp{Select new generation}
        %Compute the Pareto Front form the archive $A$\;
        $n_{\text{samples}} \leftarrow p_{\text{mut}}B + 2(1 - p_{\text{mut}})B$\;
        Sample uniformly $n_{\text{samples}}$ cells in the grid with replacement\;
        In each cells' Pareto front, sample uniformly a solution $\mathbf{x}$\;
        \texttt{\\}
        
        \tcp{Get offsprings}
        Apply $h_{\text{mut}}$ to $p_{\text{mut}}B$ of the sampled solutions\;
        Apply $h_{\text{cross}}$ to the rest of the sampled solutions\;
        The set of offsprings is the union of the outputs of both operators\;
        \texttt{\\}
        
        \tcp{Addition in the archive}
        For each offspring, find the cell corresponding to its descriptor\;
        Add offsprings to their cells and recompute Pareto fronts\;
        \texttt{\\}
        
        \tcp{Update num steps}
        $n_{\text{steps}} \leftarrow n_{\text{steps}} + 1$\;
    }
    
    \caption{\mome pseudo-code}
    \label{alg:pseudocode}
\end{algorithm}
}
\end{figure}

\section{Domains}
\label{sec:domains}
We consider two types of test domains that all exhibit large range of search space dimensions with two objectives to be maximized.\\

\textbf{Rastrigin domains.} The Rastrigin problem \cite{rastrigin1974systems} is  a benchmark optimization problem for many \qd methods \cite{fontaine2019cmame, cully2020emmiters}. We extend it to the multi-objective setting: following the initial definition of the problem, the search space is defined using $100$ dimensions. We define two objective functions, both being Rastrigin objectives but with their extrema shifted:

\begin{align}
    \begin{cases}
      f_1(\mathbf{x}) = \sum\limits_{i=1}^n [(x_i - \lambda_1)^2 - 10\cos (2\pi (x_i - \lambda_1))] \\
      f_2(\mathbf{x}) = \sum\limits_{i=1}^n [(x_i - \lambda_2)^2 - 10\cos (2\pi (x_i - \lambda_2))]
    \end{cases}       
\end{align}
where $\lambda_1 = 0.0$ and $\lambda_2 = 2.2$ correspond respectively to the optimums of $f_1$ and $f_2$.

In the first task, noted Rastrigin-multi \cite{cully2020emmiters}, the descriptors of a solution $\mathbf{x}$ are defined as the projection on the first and second dimensions in the search space and clipped:

\begin{align}
    \begin{cases}
      c_1(\mathbf{x}) = \text{clip}(x_1) \\
      c_2(\mathbf{x}) = \text{clip}(x_2) \\
    \end{cases}
    \text{where} \ 
    clip(x_i) = 
    \begin{cases}
      x_i & \text{if $-2.0 \leq x_i \leq 4.0$} \\
      -2.0 & \text{if $x_i < -2.0$} \\
      4.0 & \text{otherwise} 
    \end{cases}
\end{align}

The objective of this descriptors definition is to create several domains in the search space capable of filling the entire descriptors space, while not being equivalent in terms of maximum objective values that are reachable.

The second task we consider changes the descriptors function $\mathbf{c}$ and replace it with:
\begin{align}
    \begin{cases}
      c_1(\mathbf{x}) = \frac{1}{50}\sum\limits_{i=1}^{50} \text{clip}(x_i) \\
      c_2(\mathbf{x}) = \frac{1}{50}\sum\limits_{i=50}^{100} \text{clip}(x_i) \\
    \end{cases}
\end{align}
corresponding to the averages of the first and the second half of the components of $\mathbf{x}$. This task, inspired from \cmame \cite{fontaine2019cmame} (we use a different clipping function), is called Rastrigin-proj as it creates a unimodal structure in the objectives spaces in opposite to the Rastrigin-multi formulation where multiple local minima exist.\\

\begin{figure}
    \centering
    \includegraphics[width=0.35\textwidth]{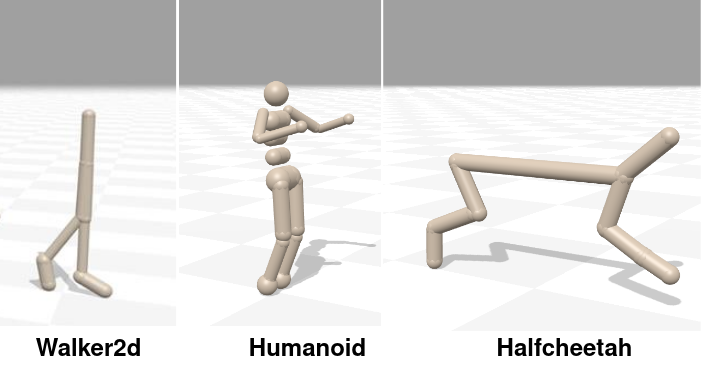}
    \caption{The Brax domain tasks in which the agent must learn to run forward as fast possible while minimizing the control cost. The descriptor space is defined as the contact between the robot legs and the ground frequency \cite{cully2015robots, colas2020scaling}.}
    \label{fig:brax_envs}
\end{figure}

\textbf{Brax domains.} The last set of tasks we study is based on the Brax suite \cite{freeman2021brax}, a Physics Engine for Large Scale Rigid Body Simulation written in Jax. In all the tasks we consider, our goal is to find the parameters of a feed-forward neural network so as to control a biped robot to run forward as fast as possible while minimizing the control cost. 

Formally, we define an observation space $\mathcal{O}$, an action space $\mathcal{A}$ and a controller $\pi_{\theta}: \mathcal{O} \rightarrow \mathcal{A}$ implemented by a neural network and parametrized by $\theta \in \Theta$ where $\Theta$ is here the search space. An agent, in our case the biped robot, interacts with the environment at discretized time intervals. At each time step $t$, the agent receives an observation $\mathbf{o}_t$ that comprises statistics about the current agent state and its environment and uses its controller to take an action $\mathbf{a}_t$ that corresponds to torques applied to the robot joints. The two function objectives are respectively defined as the sum of the control cost and forward velocity over $T=1000$ timesteps:

\begin{align}
\label{eq:brax_obj}
    \begin{cases}
      f_{1}(\theta) = - \sum\limits_{t=1}^T ||\mathbf{a}_{t}||_2 \\
       f_{2}(\theta) = \sum\limits_{t=1}^T \frac{x_{t} - x_{t-1}}{\Delta t} 
    \end{cases}
\end{align}

where $||.||_2$ is the Euclidean norm and $x_t$ is the robot center's of gravity position along the forward axis at time $t$ and $\Delta t$ is the time step.

The descriptors are computed as the mean contact frequency between the robot's two legs and the ground in order for the agent to learn to run as fast as possible with diversity in the way it moves \cite{cully2015robots, fontaine2021differentiable}:

\begin{align}
      c_{i}(\theta) = - \frac{1}{T }\sum\limits_{t=1}^T c_{i, t}, \ i=1,2
\end{align}
where $c_{i, t} = 1$ if leg $i$ is in contact with the ground at time $t$ and $0$ otherwise.

We consider three tasks in this domains that respectively correspond to three robot morphologies displayed on Figure~\ref{fig:brax_envs}, namely HalfCheetah, Walker2d and Humanoid. Note that the objectives as defined in Equations~\ref{eq:brax_obj} correspond strictly to the ones used for HalfCheetah. In Walker2d and Humanoid, Brax authors weigh the control cost by different factors and add bonuses to the forward velocity term if the robot satisfy some constraints such as keeping its center of gravity high enough. We re-use the exact same coefficients and terms in our experiments. For each task we consider a controller implemented by a two layers of $128$ neurons feed-forward neural network, the difference of observation and action size make vary slightly the search space size between tasks, see Table~\ref{tab:brax_specs} for more details. 

\begin{table}[]
\centering
\caption{Specifications of the Brax domains tasks.}
\begin{tabular}{ l | l l l }
 Task & Observation size & Action size & Search space size \\ 
 \midrule
 Halfcheetah & 23 & 6 & 21132 \\  
 Walker2d & 20 & 6 & 20748 \\  
 Humanoid & 299 & 17 & 59298 \\  
\end{tabular}
\label{tab:brax_specs}
\end{table}

\begin{figure*}[ht!]
    \centering
    \includegraphics[width=0.85\textwidth]{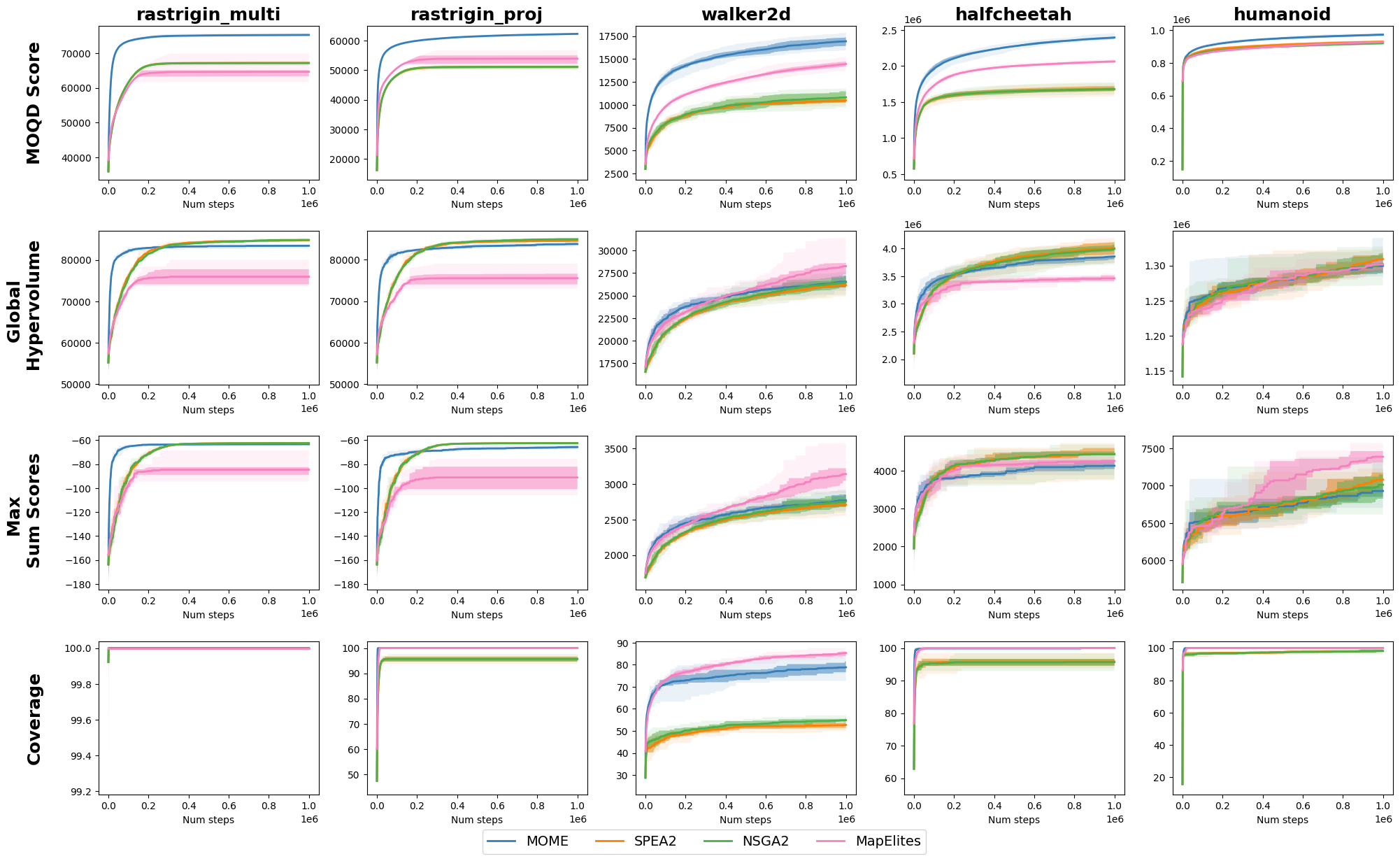}
    \caption{Comparison of \mome, \me, \nsga and \spea on the five tasks we consider. Each method for each task has been launched with $50$ different runs for the Rastrigin domains and with $20$ runs for the Brax domains. The curves show the average performance. The darker area correspond to the area between the 25\% and 75\% percentile while the lighter area shows the minimum and maximum values. As detailed in Section~\ref{sec:fair_exp}, the reported metrics are computed on a passive archive for \me, \nsga and \spea.}
    \label{fig:results_metrics}
\end{figure*}

\section{Experiments}

\begin{figure*}[ht!]
    \centering
    \includegraphics[width=0.85\textwidth]{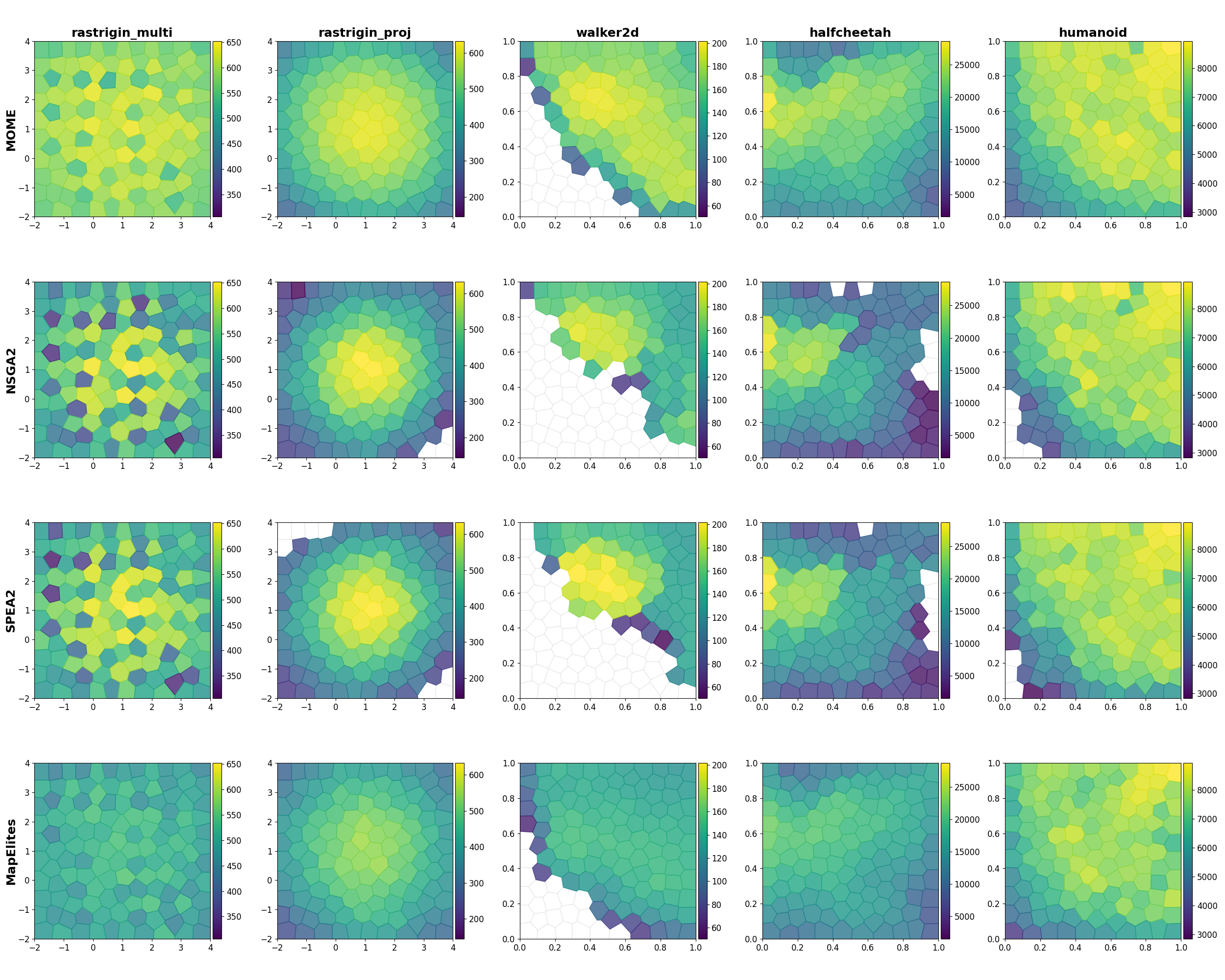}
    \caption{Results of one optimization run for \mome, \me, \nsga and \spea on the considered tasks, where the hypervolume in each cell of the grid is represented. As mentioned in Section~\ref{sec:fair_exp}, these results are obtained with a passive archive that is similar to the one used by \mome to make the results comparable. Nonetheless, we remind the reader that these archive are obtained by inserting at each generation all the generated offsprings. Plotting the same quantity but only with the final population generated by \spea and \nsga would result in more scarce and focused coverage.}
    \label{fig:results_grids}
\end{figure*}

As stated in Section~\ref{sec:moqd}, the objective of the \moqd framework is to find high-performing Pareto front in all the niches of the descriptor space. We compare \mome to two types of algorithm on the five tasks described in Section~\ref{sec:domains}. First, we compare \mome to the standard \me algorithm that is suited to optimize for descriptor space coverage but does not handle natively multi-objective optimization. Thus, in all experiments conducted with \me, the mono-objective fitness will be defined as the sum of the objectives. Second, we compare \mome to \nsga and \spea, both methods being references for multi-objective optimization but they do not seek natively for coverage in the descriptor space.

\subsection{Metrics for Fair comparisons}
\label{sec:fair_exp}

\textbf{Fair comparison with \me}. \mome and \me differ in both their objective and the tessellation method which implies that comparisons must be conducted carefully to get sensible conclusions.  
\mome maintains a Pareto front inside each cell of its grid whereas \me stores a single solution in each cell. As a consequence, if the number $d$ of cells in the Tessellation is the same for both methods, then \mome would maintain an archive up to $P$ times larger than \me where $P$ is the maximum size of the Pareto fronts. This large difference in archive size bias the comparison between both methods. Thus, when comparing both methods, we use a Tessellation of $M \times P$ cells for \me. However, this difference in the number of cells invalidates the comparison of some metrics such as the \moqd score or the grid coverage. Therefore, we introduce a passive archive for \me. The passive archive is the exact same archive than the one used for \mome, but we refer to it as passive because it is used only for metrics tracking, all the solutions produced by \me are introduced in this archive with the same addition rules used in \mome but solutions are never sampled into it which ensures that the introduction of this archive does not influence \me behavior. All the metrics reported for \me and compared to \mome are computed from the passive archive. Another element to take into consideration when running this comparison comes from the fact that while \me is designed to cover the descriptor space as \mome, it optimizes for the sum of objective and thus can not be expect to match native multi-objective optimization methods. Nonetheless, we report the \moqd score, the hyper-volume of the global Pareto front and the maximum sum of objectives found by both \mome and \me.\\

\textbf{Fair comparison with multi-objective methods}. Both \nsga and \spea are reference methods for multi-objective optimization. However none of them is designed to cover a descriptor space as \mome does. As a matter of fact, while these methods maintain a population they do not maintain a tessellation over the descriptor space which we could use to monitor their \moqd score and descriptor space coverage to compare with \mome. To make this comparison possible, we ensure that the population size used is the same as for \mome and as for \me and we introduce a passive archive with the exact same characteristics than the one used for \mome but that is used for monitoring purpose only and does not affect in any way \spea and \nsga behaviors.  In addition, as both methods are designed to find a single high-performing Pareto front instead of a collection, we also report the hyper-volume of the global Pareto front, dubbed global hyper-volume, found by \mome and compare it to the ones built by \nsga and \spea. \\

\textbf{Fair comparison as evolutionary methods.}
All the methods we consider are based on evolution. As stated above, we ensure that the archive sizes as well as the batch sizes used are the same for all methods. To ensure fair comparison, we also use the exact same mutation and crossover operators, with the same hyper-parameters and in the same proportions, for all the considered algorithms.\\

\textbf{Design of Experiments}
For Rastrigin domains, we use the polynomial mutator \cite{deb2007self} with $\eta=1$ and a standard crossover and keep the mutation/crossover ratio identical for every method. For Brax domains, we use the Iso+LineDD crossover operator \cite{vassiliades2018discovering} with the same parameters $\sigma_1=0.005$ and $\sigma_2=0.05$ in every experiment. For every experiment, we use a CVT grid \cite{vassiliades2016scaling} of size $128$ with a maximum number of $50$ solutions in each niche. Hence, \me is run using a grid of $6400$ cells and \nsga and \spea are run using a population size of $6400$ solutions. We ran every experiment for a total of $1,000,000$ evaluations, and ran $20$ runs on the Brax domains and $50$ runs on the Rastrigin domains.

\subsection{Results and Discussion}

The main results are displayed on Figure~\ref{fig:results_metrics} and Figure~\ref{fig:results_grids}. The main conclusion of the experiments is that \mome is indeed able to outperform traditional approaches for a multi-objective quality diversity purpose. We use a Wilcoxon signed-rank test \cite{wilcoxon1992individual} to compare the distribution of the scores obtained over different seeds where the null hypothesis is that the obtained scores have the same median. It shows that \mome constantly outperforms other approaches in the MOQD score with p-values lower than $0.003$ in every experiment. This confirms that \mome is better at finding diverse and high-performing solutions (in the sense of Pareto efficiency) than traditional approaches. It can clearly be seen on Figure~\ref{fig:results_grids} where one can see that \nsga and \spea find high performing niche and keep improving only those niches while \mome keeps improving good quality solutions spanning the descriptor space.

In every domain, \mome manages to explore a diverse set of solutions. A notable result is the difference between \textit{rastrigin\_multi} and \textit{rastrigin\_proj} where the only difference is the descriptors: in \textit{rastrigin\_multi}, every method manages to span the whole descriptor space even though \nsga and \spea do not improve some niches after initialization. On the other hand, in \textit{rastrigin\_proj}, both methods fail to produce any "extreme" solution in the descriptor space. It is also interesting in the \textit{walker2d} and \textit{humanoid} experiments where \mome explores parts of the descriptor space that were not covered at initialization whereas \nsga and \spea never do. 

\begin{figure}
    \centering
    \includegraphics[width=0.4\textwidth]{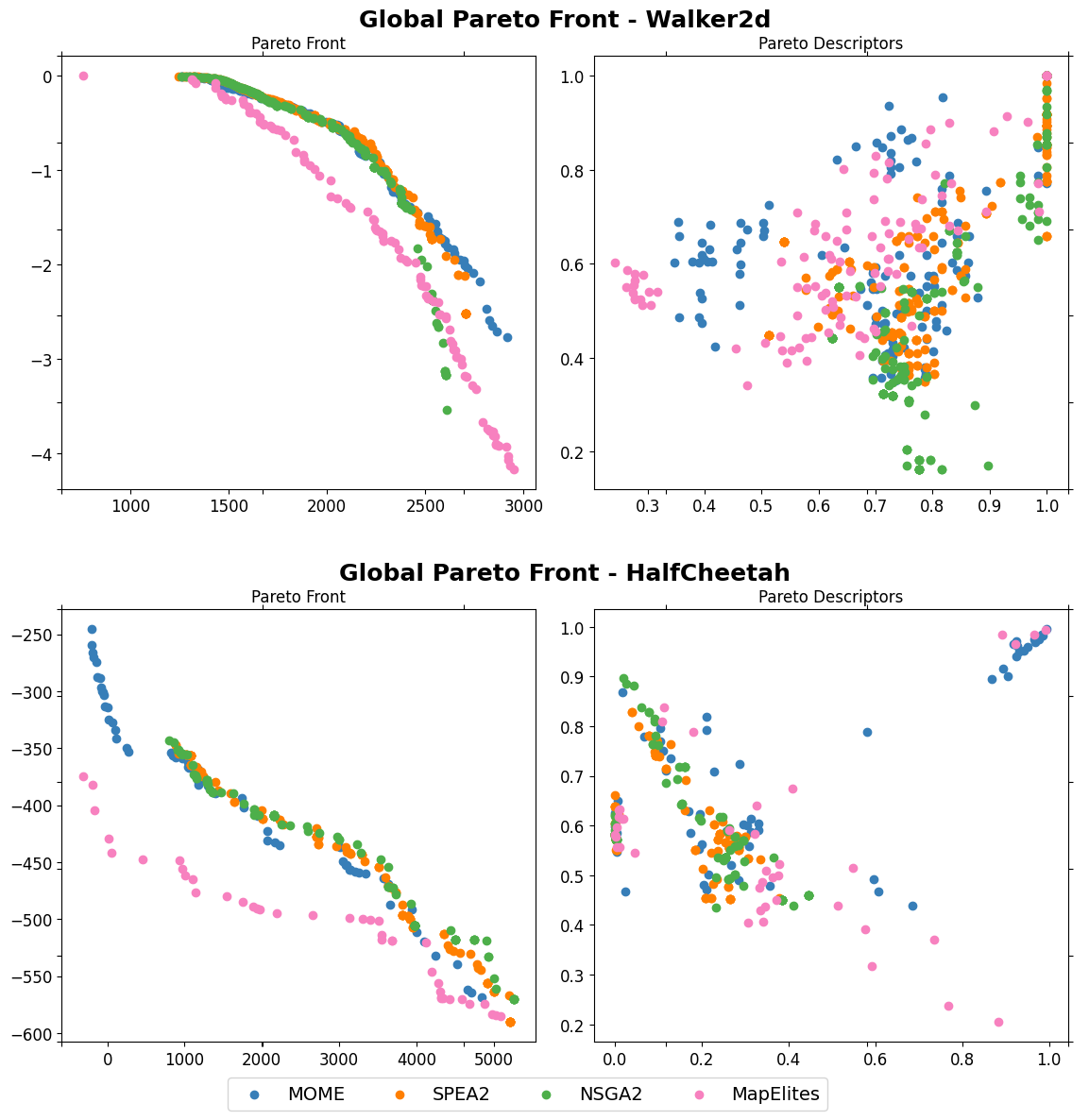}
    \caption{Pareto Fronts extracted by the four methods on Half Cheetah and the Walker2d experiments.}
    \label{fig:gpf_hc}
\end{figure}

Moreover, the global Pareto Front found by \mome is competitive with the ones found by other methods. Indeed, while on toy Rastrigin experiments, the global Pareto front is consistently slightly worse, on Brax experiments, a Wilcoxon ranked-test could not highlight a statistically significant difference between the three methods in terms of hypervolume of the global Pareto Front (p-values greater than $0.38$, $0.08$ and $0.21$ for Walker2d, HalfCheetah and Humanoid respectively). It is a significant result as it means that for the same number of evaluations, \mome finds a more diverse set of candidates with a high quality for the best performing solutions: for instance, in the Pareto Fronts extract on Walker2d represented on Figure~{\ref{fig:gpf_hc}}, the solutions found by \mome are well spread over the descriptor space while \nsga and \spea are stuck to some niches only. Even more interesting is the fact that \mome sometimes manages to extract solutions in areas of the Pareto Front unexplored by \nsga and \spea: for instance, on HalfCheetah, \mome manages to extract solutions corresponding to low control cost and low forward speed (upper left part of the Pareto Front) while \spea and \nsga only consider solutions with high control cost and high forward speed (lower right part of the Pareto Front).

We voluntarily did not normalize the Pareto Fronts to keep the original scales of those domains as it should be noted that the order induced by the hypervolume indicator is independent to scale. One can notice that \me performs very well on the different Brax domains, especially for the single objective (max sum scores) metric. In fact, it even competes with multi-objective approaches on the global hypervolume metric as we chose to use the true attainable minima for the reference point. Indeed, for \textit{walker2d} for instance, the forward velocity takes its values between $0$ and $4000$ while the control cost experimentally varies between $0$ and $-4$ but we used $(0, -10)$ as a reference point. Hence, the forward velocity accounts for most of the hypervolume of the Pareto Front. On the other hand, on Rastrigin domains where both scores have the same spread, \me clearly underperforms in terms of hypervolume. 

An important benefit of \mome is that the intrinsic diversity implied by the grid is very helpful for the final decision maker. Indeed, in most multi-objective optimization problem, after the Pareto Front is obtained, a final decision is generally made by an expert. Introducing diversity in the descriptor space allows to retrieve solutions with lower quality (dominated by others) but potentially with better properties, such as robustness, a property already observed for \me \cite{cully2015robots}. It should also be noted that our results are not comparable with state of the art methods using policy gradient \cite{nilsson2021policy, pierrot2021diversity}. Indeed, incorporating policy gradient for multi-objective optimization is not straightforward and is a future direction of our work.

\section{Conclusion and Future Works}

In this work, we defined the Multi Objective Quality Diversity (\moqd) framework, which we believe can frame applications in robotics but also others such as protein design. We introduced \mome, an extension of \me to multi-objective optimization. Our experimental evaluation shows very promising results as \mome outperform traditional multi-objective approaches on the diversity metric and even comes close to their performance on the quality of the global Pareto Front. We emphasize that the diversity brought by \mome is crucial for decision making where the highest performing solution is not always the best, as some properties are not possible to translate into objectives. We plan in our future works to extend our MOQD framework to improve MOME sample efficiency with techniques using gradient information inspired from \pgame, \cmame and differentiable QD \cite{nilsson2021policy, fontaine2019cmame, fontaine2021differentiable}.

\newpage

\bibliographystyle{ACM-Reference-Format}
\bibliography{main}

\end{document}